\documentclass{article}
\usepackage{spconf,amsmath,graphicx}
\usepackage{subcaption}
\usepackage{float} % Allows [H] placement

\usepackage{enumitem}
\usepackage{booktabs} % For better table lines
\setlist{nosep, leftmargin=14pt}

\title{Synthesizing Proton-Density Fat Fraction and $R_2^*$ from 2-point Dixon MRI with Generative Machine Learning}

\twoauthors
 {Suma Anand \sthanks{Work done as an intern at insitro.}}
	{University of California, Berkeley\\
	Electrical Engineering and Computer Sciences\\
	Berkeley, CA, USA}
 {Kaiwen Xu, Colm O'Dushlaine, Sumit Mukherjee}
	{insitro, inc. \\
	South San Francisco, CA, USA}
\begin{document}
%\ninept
%
\maketitle
% 100 -150 word abstract; currently 129
\begin{abstract}
Magnetic Resonance Imaging (MRI) is the gold standard for measuring fat and iron content non-invasively in the body via measures known as Proton Density Fat Fraction (PDFF) and $R_2^*$, respectively. However, conventional PDFF and $R_2^*$ quantification methods operate on MR images voxel-wise and require at least three measurements to estimate three quantities: water, fat, and $R_2^*$. Alternatively, the two-point Dixon MRI protocol is widely used and fast because it acquires only two measurements; however, these cannot be used to estimate three quantities voxel-wise. Leveraging the fact that neighboring voxels have similar values, we propose using a generative machine learning approach to learn PDFF and $R_2^*$ from Dixon MRI. We use paired Dixon-IDEAL data from UK Biobank in the liver and a Pix2Pix conditional GAN\cite{Isola_2017_CVPR} to demonstrate the first large-scale $R_2^*$ imputation from two-point Dixon MRIs. Using our proposed approach, we synthesize PDFF and $R_2^*$ maps that show significantly greater correlation with ground-truth than conventional voxel-wise baselines.
\end{abstract}
\section{Introduction}
\label{sec:intro}
Magnetic Resonance Imaging (MRI) is a premier non-invasive imaging modality, not only for visualizing soft tissues but also estimating quantitative indicators of fat  and iron. Proton Density Fat Fraction (PDFF) and $R_2^*$ (a relaxation time constant) are two such indicators. They are currently the clinical gold standard for diagnosing metabolic dysfunction associated steatotic liver disease (MASLD) \cite{gu2019diagnostic} and iron overload in the liver \cite{rostoker2019histological}. Conventionally, PDFF and $R_2^*$ are estimated voxel-wise by taking measurements at multiple time points and recovering three underlying quantities: water, fat, and $R_2^*$. However, conventional techniques require at least three measurements for this estimation. Alternatively, Dixon MRI \cite{dixon1984simple} is a widespread technique that acquires only two measurements. It is possible to estimate fat and water from these measurements; however, they are known to be confounded by other factors such as $R_2^*$ and $T_1$ relaxation, leading to underestimated fat fraction when PDFF $< 50 \%$ \cite{reeder2012proton, bydder2008relaxation}. Moreover, it is impossible to recover $R_2^*$ accurately from Dixon using conventional techniques because there are only two measurements. %Fat fraction estimated from Dixon without accounting for these factors is referred to as Signal Fat Fraction (SFF) or simply Fat Fraction (FF) and known to be less accurate than PDFF \cite{reeder2012proton}.

A more recent and accurate MRI technique known as IDEAL acquires six measurements \cite{reeder2004multicoil}. PDFF and $R_2^*$ are estimated from these measurements via a voxel-wise regularized nonlinear least squares approach \cite{bydder2020constraints}. However, IDEAL is significantly slower than Dixon and often acquired only on a single slice. Additionally, the voxel-wise estimation approach does not leverage similarity between neighboring pixels. However, in practice, radiologists often look at averaged PDFF and $R_2^*$ over a region of interest in the liver \cite{wood2005mri, caussy2018noninvasive}. This suggests that neighboring voxels are similar, and thus methods such as convolutional neural networks can leverage this local structure to estimate PDFF and $R_2^*$ with fewer than three measurements. With this in mind, we propose a CNN-based generative machine learning approach to recover PDFF and $R_2^*$ from Dixon MRI. While there exist some machine learning-based methods to estimate PDFF from Dixon \cite{wang2023deep}, and others to estimate $R_2^*$ from MRI data with greater than 2 measurements \cite{gao2021accelerating}, $R_2^*$ estimation from two-point Dixon has not been explored. We focus on the liver region using publicly available data in UK Biobank \cite{littlejohns2020uk}. Using a Pix2Pix model \cite{Isola_2017_CVPR}, we demonstrate a proof-of-concept approach that i) is able to estimate $R_2^*$ with good accuracy where the voxel-wise baseline totally fails and ii) can estimate PDFF with greater accuracy than conventional Dixon.

\section{Background}

MRI-based methods for measuring fat exploit a property known as chemical shift, where water and fat resonate at different frequencies \cite{nishimura1996principles}; thus, imaging measurements are made when water and fat have different phase offsets from one another.

In Dixon MRI, two images are acquired when water and fat are in-phase and out of phase at times $t_1$ and $t_2$.They can be characterized by two simple equations \cite{dixon1984simple}:
\begin{align}
    S(t_1) = W + F \\
    S(t_2) = W - F
\end{align}
where $S(t_i)$ are the acquired image magnitudes, $W$ refers to a voxel-wise map of water content and $F$ a voxel-wise map of fat content.

These measurements can be added and subtracted, respectively, to estimate the underlying water and fat:
\begin{align}
    W = 0.5*(S(t_1) + S(t_2)) \\
    F = 0.5*(S(t_1) - S(t_2))
\end{align}
These equations are simplistic and do not account for $R_2^*$ relaxation or multiple fat peaks. A more complex signal model is given by \cite{hernando2013multipeak}:
\begin{align}
    S(t_1) = (W + F)e^{-t_1R_2^*} \\
    S(t_2) = (W - F)e^{-t_2R_2^*} 
\end{align}
However, because there are three underlying quantities and two measurements, $W$, $F$, and $R_2^*$ cannot be recovered by conventional methods. 

More complex signal models that incorporate additional signal dynamics can lead to accurate water and fat estimation when used with multiple ($>2$) measurements. For instance, fat contains multiple different chemical species that resonate at different frequencies. A model that accounts for these resonances is known as a  multi-peak model \cite{yu2008multiecho} and is given by:

\begin{align}
S(t_1) &= \left( W + \sum_{i=1}^N F_i e^{-j 2 \pi f_i t_1} \right) e^{-t_1 R_2^*} \\
S(t_2) &= \left( W + \sum_{i=1}^N F_i e^{-j 2 \pi f_i t_2} \right) e^{-t_2 R_2^*} \\
&\vdots \nonumber \\
S(t_M) &= \left( W + \sum_{i=1}^N F_i e^{-j 2 \pi f_i t_M} \right) e^{-t_M R_2^*}
\end{align}
 where $j=\sqrt{-1}$, $F_i$ refers to each species of fat with corresponding resonant frequency $f_i$, and there are N total species and M measurements. IDEAL \cite{reeder2004multicoil} is a method in which 6 measurements are acquired, and a multi-peak signal model is used to estimate PDFF and $R_2^*$ via a regularized least squares algorithm \cite{bydder2020constraints}. This method is currently the clinical gold standard for accurate PDFF computation \cite{reeder2012proton}.

\label{sec:background}

\section{Methods}
\label{sec:methods}
\subsection{Data Preprocessing}
All data was from UK Biobank \cite{littlejohns2020uk}, in which there are more than 30,000 Dixon and IDEAL scans acquired in the same session (i.e., more than 30,000 unique subjects). Dixon data was acquired with a field of view from neck to knee, while IDEAL data is single-slice with a matrix size of $232 \times 256$ and $1.7 \times 1.7  \times 10$mm resolution. Dixon data was first pre-processed using the pipeline developed by Liu et al \cite{liu2020genetic}: resampled to 3mm slice thickness, in-plane matrix size of $224 \times 174$, and corrected for water-fat swaps \cite{basty2023artifact, liu2020genetic}. Because the IDEAL data were acquired as a single slice at the liver, the Dixon data were first resampled to 10mm slice thickness, registered to the IDEAL image, and linearly interpolated to the same matrix size and resolution as IDEAL using SimpleITK \cite{lowekamp2013design}. Four channels were extracted from the Dixon data after processing: in-phase, opposed-phase, water, and fat. The baseline fat fraction was computed based on the registered fat and water Dixon channels. Additionally, liver masks estimated from Dixon \cite{graf2024totalvibesegmentator} were extracted, resampled, and registered to IDEAL in the same way for evaluation. Mutual information (MI) between registered Dixon and IDEAL was computed to gauge the success of the registration. 

Ground-truth PDFF and $R_2^*$ were estimated using regularized non-linear least squares \cite{bydder2020constraints, liu2020genetic}. Baseline $R_2^*$ values were estimated by computing a voxel-wise exponential fit using the two Dixon timepoints. We note that this baseline is very ill-conditioned and inaccurate; however, it is the only existing method to estimate $R_2^*$ from two-point Dixon MRI. 6,976 random, anonymized paired samples were obtained. The samples with the bottom $10 \%$ of MI were discarded to ensure good quality. Data pairs with low MI were empirically found to have errors such as mismatch between the Dixon and IDEAL slice prescriptions.

\begin{figure*}[h]
  \centering
  \includegraphics[scale=0.25]{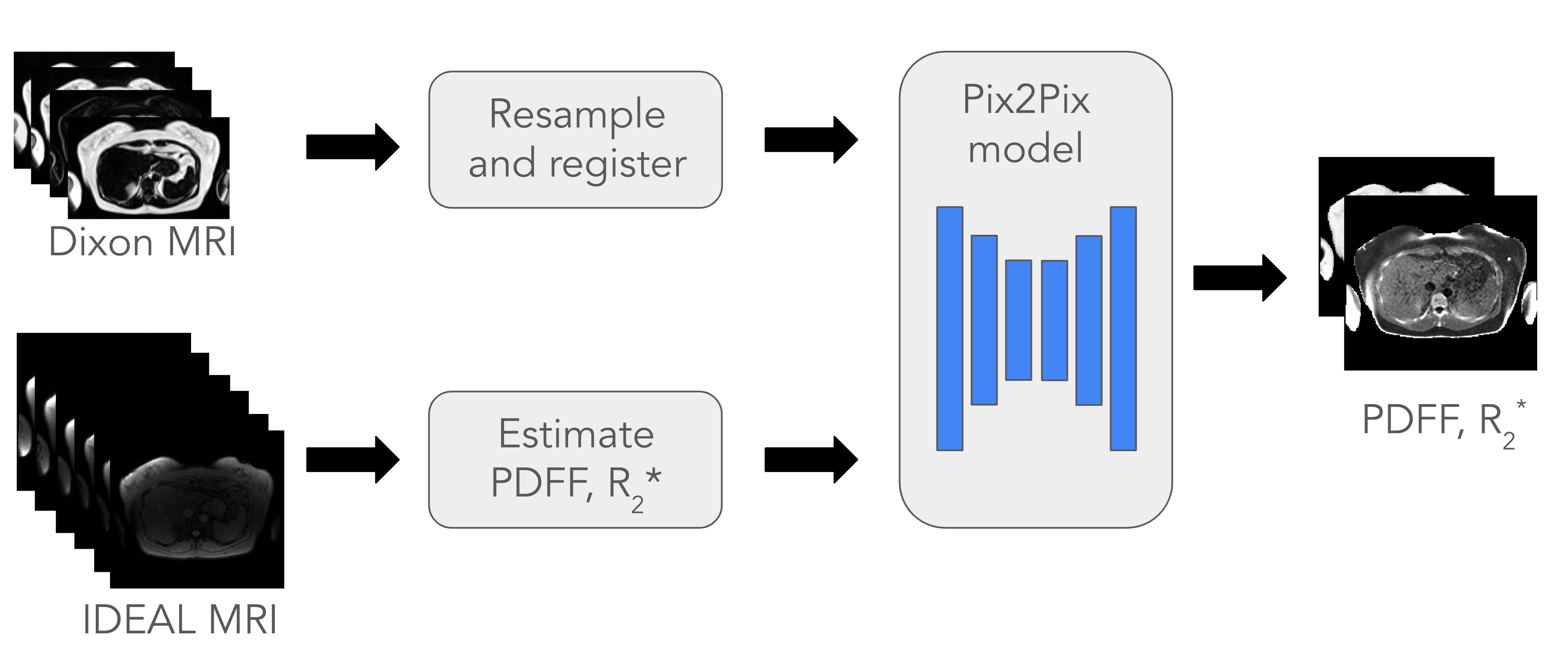}
  \caption{Pipeline for training the model to impute PDFF and $R_2^*$. Four channels of Dixon data (in-phase, opposed-phase, water, and fat) were resampled and registered to IDEAL single-slice liver images. Ground-truth PDFF and $R_2^*$ were estimated from IDEAL via nonlinear least squares. A Pix2Pix model was trained to impute PDFF and $R_2^*$ from Dixon.}
  \label{fig:example}
\end{figure*}

\subsection{Model training}
The data was split 70/20/10 into training, validation, and test, with 4,394 training, 1,255 validation samples, and 629 test samples. A Pix2Pix conditional GAN model was used \cite{Isola_2017_CVPR}, with Least Squares GAN, a reconstruction loss weight $\lambda_{L1} = 25$, and a batch size of 2.

Two model variants were trained: one that outputs both PDFF and $R_2^*$ as separate channels simultaneously, and another with separate models for PDFF and $R_2^*$ output. We refer to these as ``multi-task" and ``single-task" models, respectively. The rationale behind this is that $R_2^*$ estimation is a harder task than PDFF estimation. Fat fraction is obtained by taking a simple linear combination of the inputs, while $R_2^*$ is a nonlinear function. Thus, the network loss could be dominated by the effects of $R_2^*$ rather than PDFF in the multi-task case. In both cases, the inputs were the four Dixon channels (in-phase, opposed-phase, water, and fat). As the model was trained, $L_1$ reconstruction loss was calculated every 5 epochs on the validation set. The model was trained for 500 epochs, but the epoch that minimized reconstruction loss on the validation set was chosen for inference. To evaluate the baseline and two models, mean absolute error (MAE) was computed relative to ground-truth over the entire slice and the liver region alone using the registered liver masks (Tables \ref{tab:pdff_mae}, \ref{tab:r2s_mae}). Mean PDFF and $R_2^*$ values over the slice were also calculated (Figure \ref{fig:regression}).

\section{Results}
\label{sec:results}

\subsection{Qualitative Results}
\begin{figure}[htb]
  \centering
  \includegraphics[width=\linewidth]{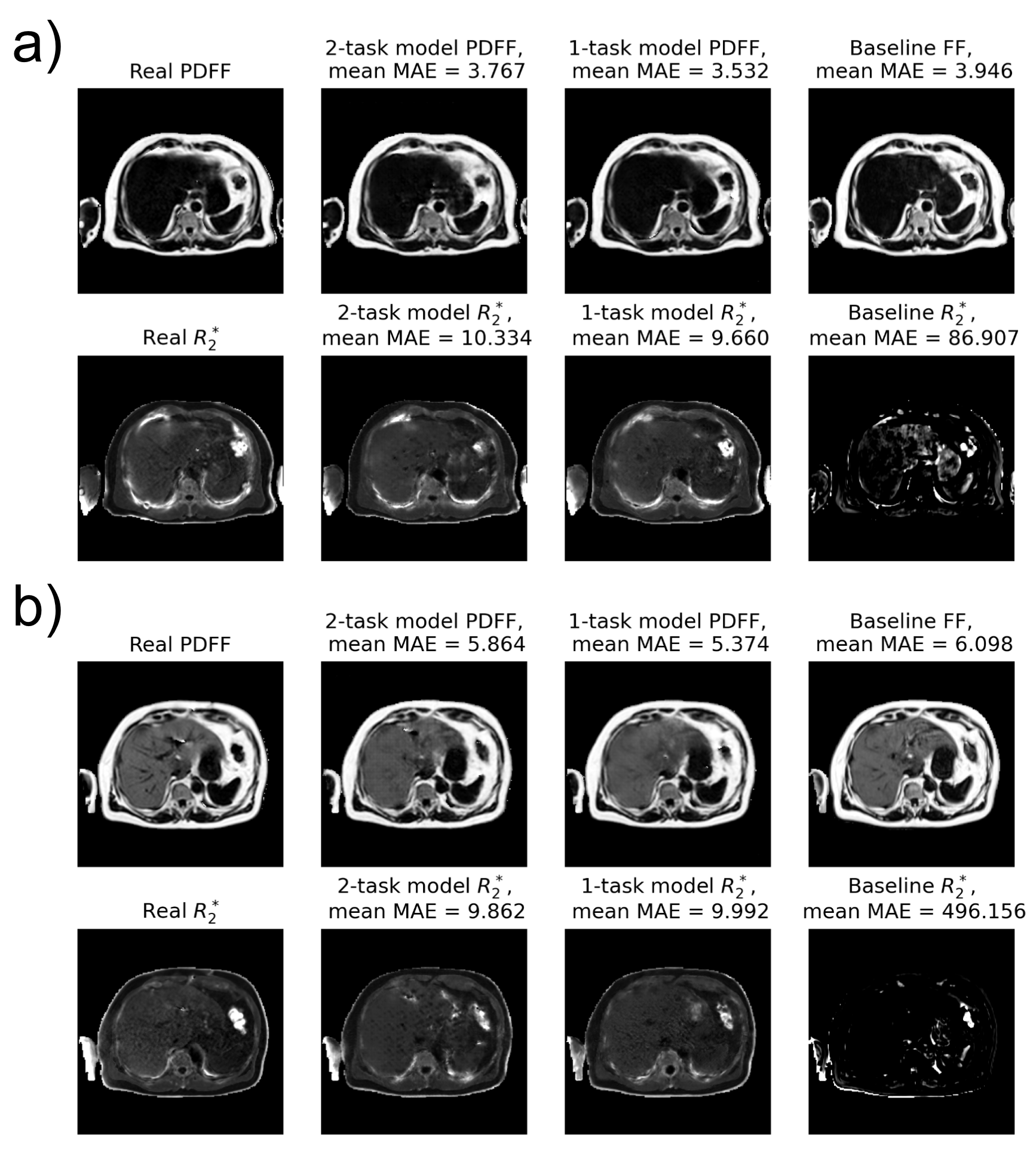}
  \caption{Qualitative results on two subjects: one with a) low ($9\%$) liver PDFF, and one with b) high ($32\%$) liver PDFF. The models produce PDFF and $R_2^*$ maps that agree qualitatively with ground-truth (left). The multi-task model has greater error in PDFF estimation in both cases, but the single-task model performs better  on the low-PDFF case than the high-PDFF case. The baseline $R_2^*$ estimation method is very inaccurate.}
  \label{fig:qualitative}
\end{figure}

Figure \ref{fig:qualitative} suggests that both model variants output PDFF and $R_2^*$ maps that visually match well with the ground truth. Figure \ref{fig:qualitative}a shows a patient with relatively low ($9\%$) average liver PDFF, while Figure \ref{fig:qualitative}b shows a patient with relatively high ($32\%$) average liver PDFF.  We find that the single-task models slightly outperform the multi-task model and show a smoother PDFF within the liver (Figure \ref{fig:qualitative}). The model results with low PDFF preserve structures in the image without loss of fine details.

\subsection{Quantitative Results}
We report average MAE over the test set for the entire image and the liver alone, excluding zero values. The single-task models have lowest MAE for PDFF and $R_2^*$ overall, while the multi-task model has the lowest MAE for $R_2^*$ in the liver, but slightly. This suggests that PDFF estimation may be confounded by estimating $R_2^*$ simultaneously. Both models outperform the baseline for PDFF and significantly for $R_2^*$. As before, we note that this is because the baseline is not at all reliable for $R_2^*$.

\begin{figure}[h]
  \centering
  \includegraphics[width=\columnwidth]{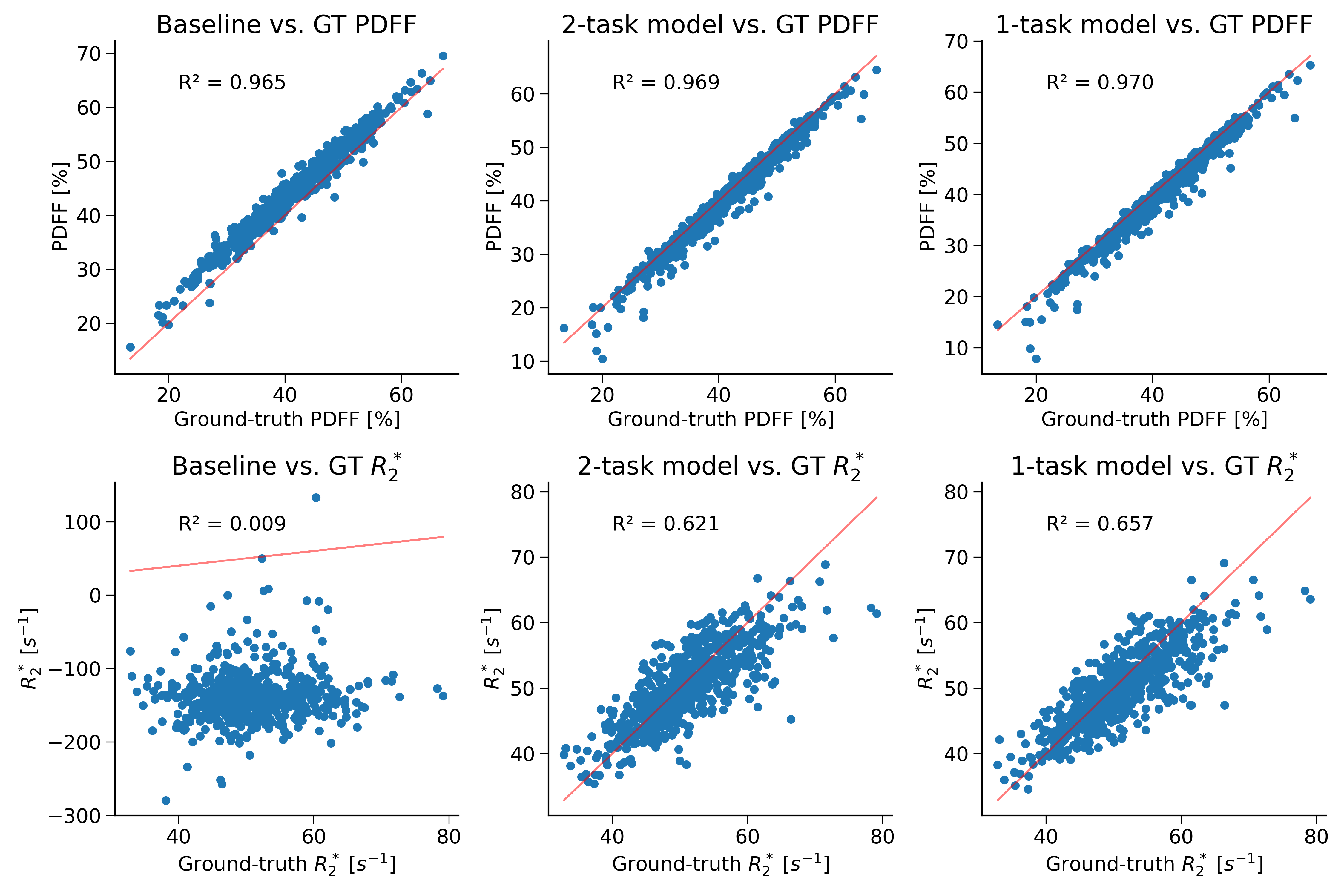}
  \caption{Mean PDFF and $R_2^*$ for baseline vs. our models and ground-truth (GT). For PDFF (top row), the baseline is less accurate overall. For $R_2^*$, the baseline shows no correlation with ground-truth, while the models show a good correlation. All methods underestimate $R_2^*$ at high values.}
  \label{fig:regression}
\end{figure}

We further investigated these results by plotting the mean PDFF and $R_2^*$ for the baseline and model variants (Figure \ref{fig:regression}, with the $y = x$ line in red. The single-task model has the highest $R^2$ value of 0.97 for PDFF, the multi-task model second-highest with 0.969 and the baseline lowest with 0.965. The baseline $R_2*$ values are totally uncorrelated with the ground-truth values ($ R^2 = 0.009$ in Figure \ref{fig:regression}). Moreover, the baseline method often predicts negative $R_2^*$ values, which are not physically possible. In contrast, the model results are strongly correlated, with $R^2$ values of 0.621 and 0.657 for the multi-task and single-task models. However, both models appear to underestimate high $R_2^*$ values. This could be due to a dearth of high $R_2^*$ data in the training set.

\begin{table}[htbp]
    \centering
    \caption{Mean Absolute Error (MAE) Comparison for PDFF}
    \begin{tabular}{lcc}
        \toprule
        Method & PDFF MAE & Liver PDFF MAE \\
        \midrule
        Baseline         & $12.68$ & $5.69$ \\
        Multi-task Model  & $9.08$ & $4.81$ \\
        \textbf{Single-task Models}   & $\mathbf{8.66}$ & $\mathbf{4.59}$ \\
        \bottomrule
    \end{tabular}
    \label{tab:pdff_mae}
\end{table}

\begin{table}[htbp]
    \centering
    \caption{Mean Absolute Error (MAE) Comparison for $R_2^*$}
    \begin{tabular}{lcc}
        \toprule
        Method & $R_2^*$ MAE & Liver $R_2^*$ MAE \\
        \midrule
        Baseline         & $216.44$ & $108.85$ \\
        Multitask Model  & $19.34$ & $\mathbf{12.23}$ \\
        Single-task Models   & $\mathbf{19}$ & $12.3$ \\
        \bottomrule
    \end{tabular}
    \label{tab:r2s_mae}
\end{table}

% \begin{table}[htbp]
%     \centering
%     \caption{Pearson correlation coefficient for PDFF and $R_2^*$}
%     \begin{tabular}{lcc}
%         \toprule
%         Method & PDFF &  $R_2^*$ \\
%         \midrule
%         Baseline         & $0.983$ & $0.096$ \\
%         Multitask Model  & $0.984$ & $0.788$ \\
%         Single-task Models   & $\mathbf{0.985}$ & $\mathbf{0.811}$ \\
%         \bottomrule
%     \end{tabular}
%     \label{tab:corr}
% \end{table}

\section{Conclusion}
This work is the first large-scale imputation of $R_2^*$ from 2-point Dixon MRI. Using two single task or multi-task Pix2Pix model variants trained to impute PDFF and $R_2^*$, we demonstrated a proof-of-concept approach that is more accurate than the Dixon-based baseline for PDFF and significantly more accurate for $R_2^*$. 

However, estimation of $R_2^*$ was less accurate than that of PDFF. We hypothesize that this is because estimating $R_2^*$ is more difficult task than estimating PDFF, because $R_2^*$ is a nonlinear function of the inputs, while PDFF is linear. This could be alleviated by inputting data from more slices to the model, adding more information. Additionally, some fine structures of the liver (e.g., vessels) may appear smoothed out. This could be mitigated by using a diffusion model approach. Training on a more diverse population (with high $R_2^*$ and high PDFF) could mitigate the underestimation of PDFF and $R_2^*$.

\section{Acknowledgments}
\label{sec:acknowledgments}
The authors would like thank the participants of the UK Biobank, whose data were used with permission. This research was conducted using the UK Biobank Resource under approved Application Number 51766.

\section{Supplementary Material}
We provide additional examples of synthesized PDFF and$R_2^*$ below.

\begin{figure*}[htbp]
    \centering
    \begin{subfigure}{0.49\textwidth}
        \includegraphics[width=\linewidth]{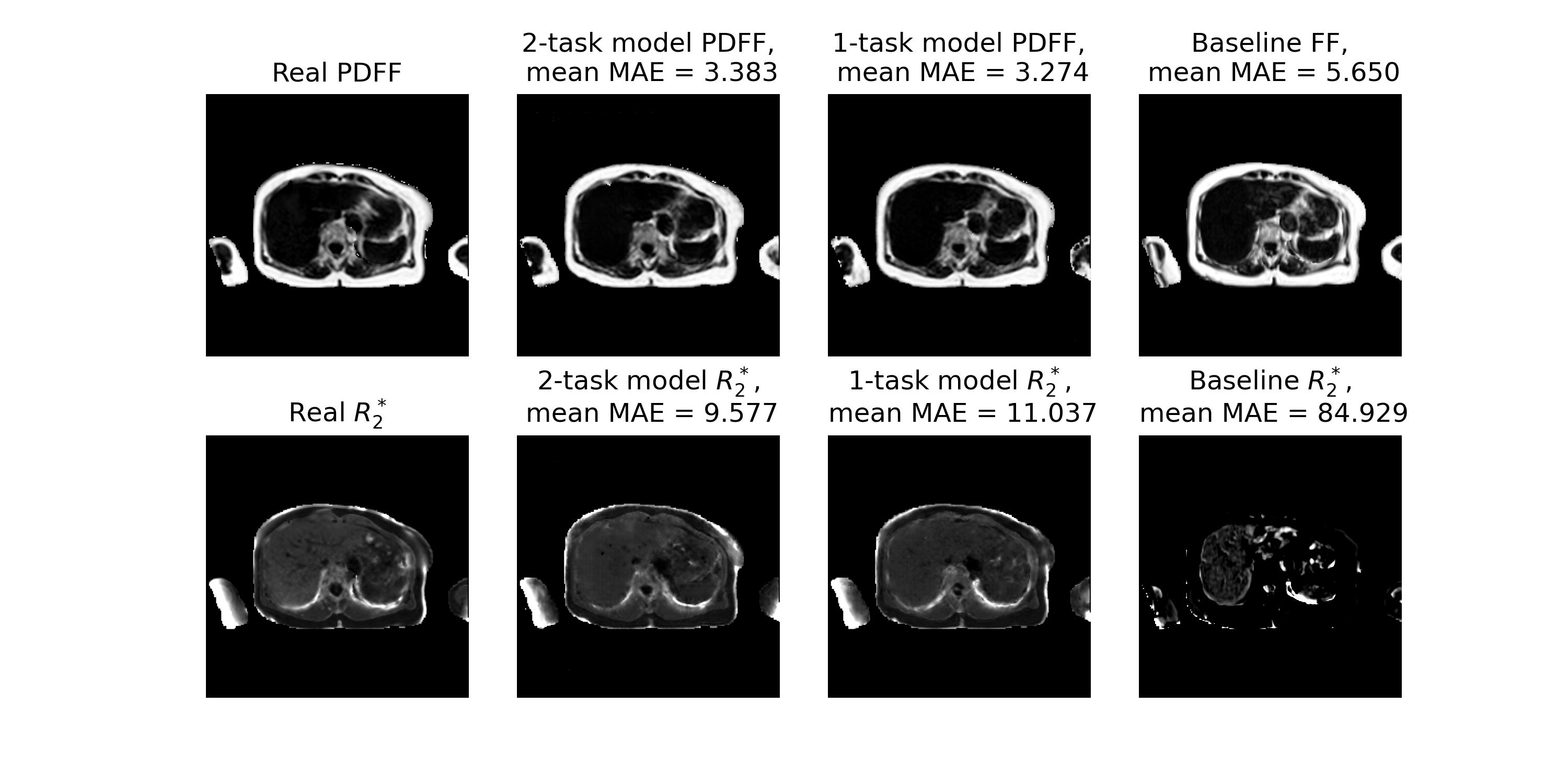}
        \subcaption{}
        \label{fig:image1}
    \end{subfigure}
    \hspace{-0.5em}
    \begin{subfigure}{0.49\textwidth}
        \includegraphics[width=\linewidth]{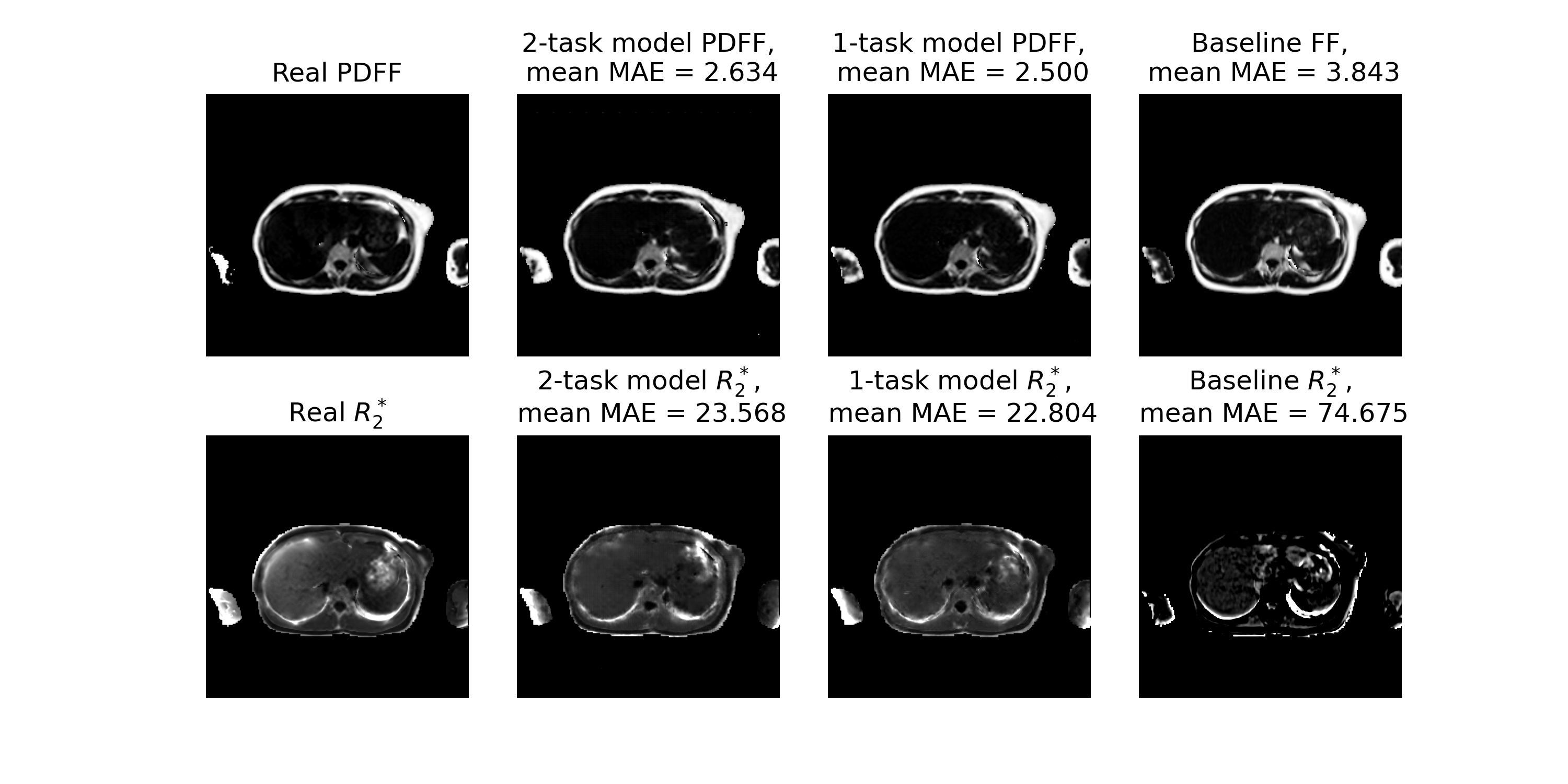}
        \subcaption{}
        \label{fig:image2}
    \end{subfigure}

    \vspace{0.05cm}

    \begin{subfigure}{0.49\textwidth}
        \includegraphics[width=\linewidth]{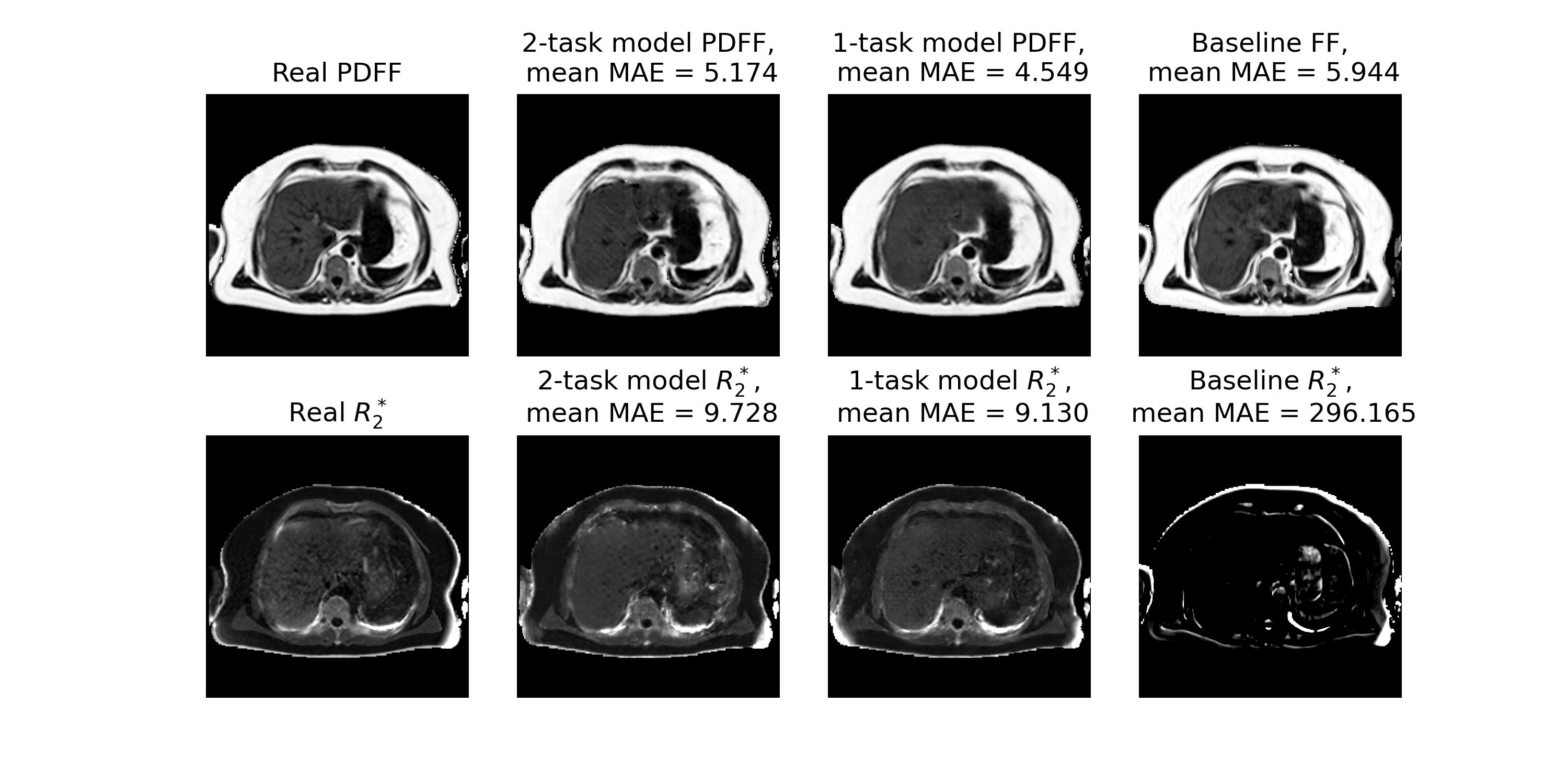}
        \subcaption{}
        \label{fig:image3}
    \end{subfigure}
    \hspace{-0.5em}
    \begin{subfigure}{0.49\textwidth}
        \includegraphics[width=\linewidth]{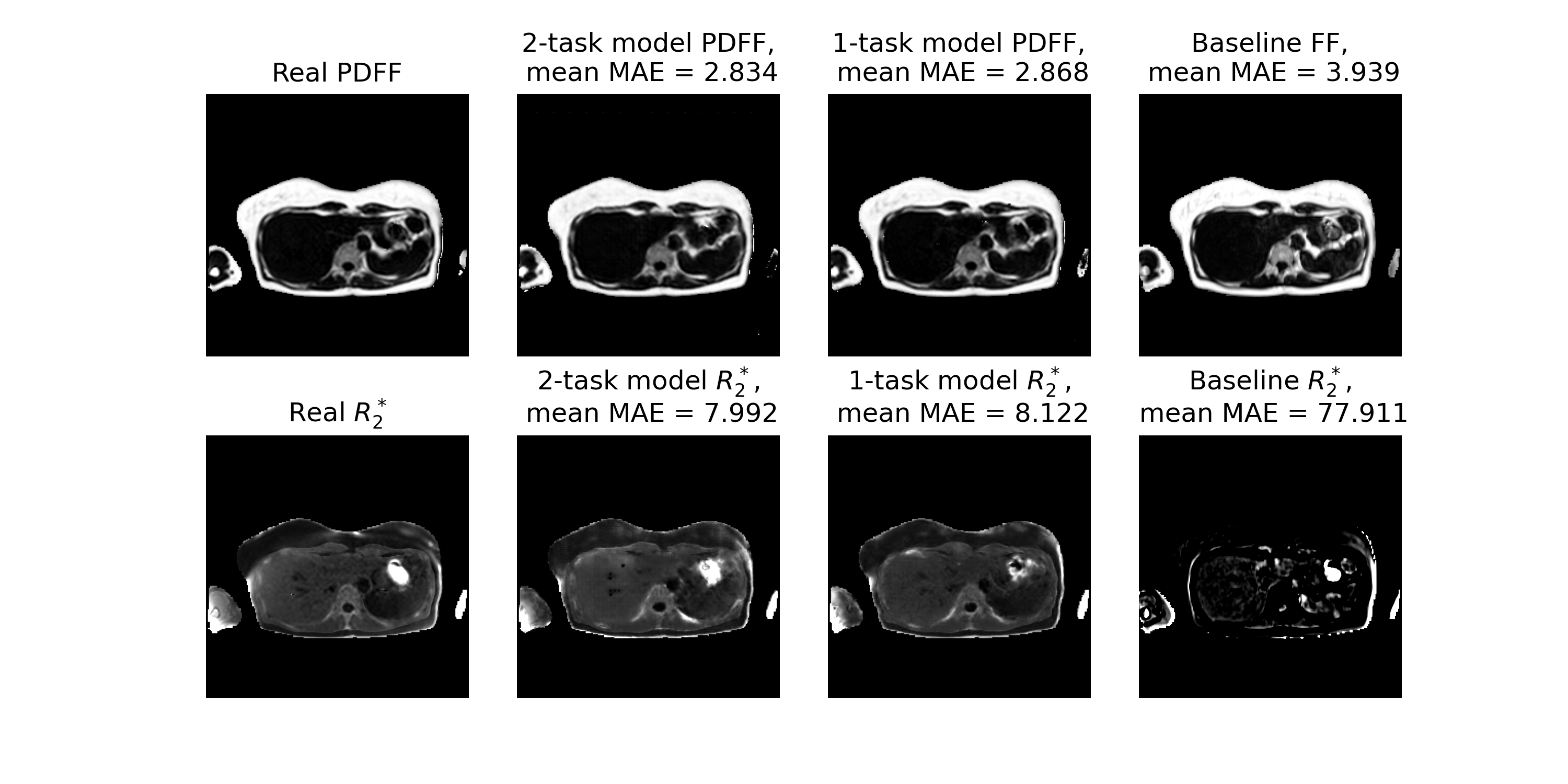}
        \subcaption{}
        \label{fig:image4}
    \end{subfigure}

    \caption{Results on four additional subjects.}
    \label{fig:four_images}
\end{figure*}

% IEEE-ISBI supports the disclosure of financial support for the project
% as well as any financial and personal relationships of the author that
% could create even the appearance of bias in the published work. The
% authors must disclose any agency or individual that provided financial
% support for the work as well as any personal or financial or
% employment relationship between any author and the sources of
% financial support for the work.

% Other types of acknowledgements can also be listed in this section.

% Reporting on real or potential conflicts of interests, or the absence
% thereof, is required in the paper. Authors are responsible for
% correctness of the statements provided in the manuscript. Examples of
% appropriate statements include:
% \begin{itemize}
%   \item ``No funding was received for conducting this study. The
%     authors have no relevant financial or non-financial interests to
%     disclose.'' 
%   \item ``This work was supported by […] (Grant numbers) and
%     […]. Author X has served on advisory boards for Company Y.'' 
%   \item ``Author X is partially funded by Y. Author Z is a Founder and
%     Director for Company C.''
% \end{itemize}

% References should be produced using the bibtex program from suitable
% BiBTeX files (here: strings, refs, manuals). The IEEEbib.bst bibliography
% style file from IEEE produces unsorted bibliography list.
% ------------------------------------------------------------------------- 
\bibliographystyle{IEEEbib}
\bibliography{strings,refs}

\begin{thebibliography}{10}

\bibitem{Isola_2017_CVPR}
Phillip Isola, Jun-Yan Zhu, Tinghui Zhou, and Alexei~A. Efros,
\newblock ``Image-to-image translation with conditional adversarial networks,''
\newblock in {\em Proceedings of the IEEE Conference on Computer Vision and Pattern Recognition (CVPR)}, July 2017.

\bibitem{gu2019diagnostic}
Jiulian Gu, Shousheng Liu, Shuixian Du, Qing Zhang, Jianhan Xiao, Quanjiang Dong, and Yongning Xin,
\newblock ``Diagnostic value of mri-pdff for hepatic steatosis in patients with non-alcoholic fatty liver disease: a meta-analysis,''
\newblock {\em European radiology}, vol. 29, pp. 3564--3573, 2019.

\bibitem{rostoker2019histological}
Guy Rostoker, Mireille Laroudie, Rapha{\"e}l Blanc, Mireille Griuncelli, Christelle Loridon, Fanny Lepeytre, Cl{\'e}mentine Rabat{\'e}, and Yves Cohen,
\newblock ``Histological scores validate the accuracy of hepatic iron load measured by signal intensity ratio and r2* relaxometry mri in dialysis patients,''
\newblock {\em Journal of Clinical Medicine}, vol. 9, no. 1, pp. 17, 2019.

\bibitem{dixon1984simple}
W~Thomas Dixon,
\newblock ``Simple proton spectroscopic imaging.,''
\newblock {\em Radiology}, vol. 153, no. 1, pp. 189--194, 1984.

\bibitem{reeder2012proton}
Scott~B Reeder, Houchun~H Hu, and Claude~B Sirlin,
\newblock ``Proton density fat-fraction: a standardized mr-based biomarker of tissue fat concentration,''
\newblock {\em Journal of magnetic resonance imaging: JMRI}, vol. 36, no. 5, pp. 1011, 2012.

\bibitem{bydder2008relaxation}
Mark Bydder, Takeshi Yokoo, Gavin Hamilton, Michael~S Middleton, Alyssa~D Chavez, Jeffrey~B Schwimmer, Joel~E Lavine, and Claude~B Sirlin,
\newblock ``Relaxation effects in the quantification of fat using gradient echo imaging,''
\newblock {\em Magnetic resonance imaging}, vol. 26, no. 3, pp. 347--359, 2008.

\bibitem{reeder2004multicoil}
Scott~B Reeder, Zhifei Wen, Huanzhou Yu, Angel~R Pineda, Garry~E Gold, Michael Markl, and Norbert~J Pelc,
\newblock ``Multicoil dixon chemical species separation with an iterative least-squares estimation method,''
\newblock {\em Magnetic Resonance in Medicine}, vol. 51, no. 1, pp. 35--45, 2004.

\bibitem{bydder2020constraints}
Mark Bydder, Vahid Ghodrati, Yu~Gao, Matthew~D Robson, Yingli Yang, and Peng Hu,
\newblock ``Constraints in estimating the proton density fat fraction,''
\newblock {\em Magnetic resonance imaging}, vol. 66, pp. 1--8, 2020.

\bibitem{wood2005mri}
John~C Wood, Cathleen Enriquez, Nilesh Ghugre, J~Michael Tyzka, Susan Carson, Marvin~D Nelson, and Thomas~D Coates,
\newblock ``Mri r2 and r2* mapping accurately estimates hepatic iron concentration in transfusion-dependent thalassemia and sickle cell disease patients,''
\newblock {\em Blood}, vol. 106, no. 4, pp. 1460--1465, 2005.

\bibitem{caussy2018noninvasive}
Cyrielle Caussy, Scott~B Reeder, Claude~B Sirlin, and Rohit Loomba,
\newblock ``Noninvasive, quantitative assessment of liver fat by mri-pdff as an endpoint in nash trials,''
\newblock {\em Hepatology}, vol. 68, no. 2, pp. 763--772, 2018.

\bibitem{wang2023deep}
Kang Wang, Guilherme~Moura Cunha, Kyle Hasenstab, Walter~C Henderson, Michael~S Middleton, Shelley~A Cole, Jason~G Umans, Tauqeer Ali, Albert Hsiao, and Claude~B Sirlin,
\newblock ``Deep learning for inference of hepatic proton density fat fraction from t1-weighted in-phase and opposed-phase mri: Retrospective analysis of population-based trial data,''
\newblock {\em American Journal of Roentgenology}, vol. 221, no. 5, pp. 620--631, 2023.

\bibitem{gao2021accelerating}
Yang Gao, Martijn Cloos, Feng Liu, Stuart Crozier, G~Bruce Pike, and Hongfu Sun,
\newblock ``Accelerating quantitative susceptibility and r2* mapping using incoherent undersampling and deep neural network reconstruction,''
\newblock {\em Neuroimage}, vol. 240, pp. 118404, 2021.

\bibitem{littlejohns2020uk}
Thomas~J Littlejohns, Jo~Holliday, Lorna~M Gibson, Steve Garratt, Niels Oesingmann, Fidel Alfaro-Almagro, Jimmy~D Bell, Chris Boultwood, Rory Collins, Megan~C Conroy, et~al.,
\newblock ``The uk biobank imaging enhancement of 100,000 participants: rationale, data collection, management and future directions,''
\newblock {\em Nature communications}, vol. 11, no. 1, pp. 2624, 2020.

\bibitem{nishimura1996principles}
D.G. Nishimura,
\newblock {\em Principles of Magnetic Resonance Imaging},
\newblock Stanford University, 1996.

\bibitem{hernando2013multipeak}
Diego Hernando, J~Harald Kramer, and Scott~B Reeder,
\newblock ``Multipeak fat-corrected complex r2* relaxometry: theory, optimization, and clinical validation,''
\newblock {\em Magnetic resonance in medicine}, vol. 70, no. 5, pp. 1319--1331, 2013.

\bibitem{yu2008multiecho}
Huanzhou Yu, Ann Shimakawa, Charles~A McKenzie, Ethan Brodsky, Jean~H Brittain, and Scott~B Reeder,
\newblock ``Multiecho water-fat separation and simultaneous r estimation with multifrequency fat spectrum modeling,''
\newblock {\em Magnetic Resonance in Medicine}, vol. 60, no. 5, pp. 1122--1134, 2008.

\bibitem{liu2020genetic}
Yi~Liu, Nicolas Basty, Brandon Whitcher, Jimmy~D Bell, Elena Sorokin, Nick van Bruggen, E~Louise Thomas, and Madeleine Cule,
\newblock ``Genetic architecture of 11 abdominal organ traits derived from abdominal mri using deep learning,''
\newblock {\em bioRxiv}, pp. 2020--07, 2020.

\bibitem{basty2023artifact}
Nicolas Basty, Marjola Thanaj, Madeleine Cule, Elena~P Sorokin, Yi~Liu, E~Louise Thomas, Jimmy~D Bell, and Brandon Whitcher,
\newblock ``Artifact-free fat-water separation in dixon mri using deep learning,''
\newblock {\em Journal of Big Data}, vol. 10, no. 1, pp. 4, 2023.

\bibitem{lowekamp2013design}
Bradley~C Lowekamp, David~T Chen, Luis Ib{\'a}{\~n}ez, and Daniel Blezek,
\newblock ``The design of simpleitk,''
\newblock {\em Frontiers in neuroinformatics}, vol. 7, pp. 45, 2013.

\bibitem{graf2024totalvibesegmentator}
Robert Graf, Paul-S{\"o}ren Platzek, Evamaria~Olga Riedel, Constanze Ramsch{\"u}tz, Sophie Starck, Hendrik~Kristian M{\"o}ller, Matan Atad, Henry V{\"o}lzke, Robin B{\"u}low, Carsten~Oliver Schmidt, et~al.,
\newblock ``Totalvibesegmentator: Full torso segmentation for the nako and uk biobank in volumetric interpolated breath-hold examination body images,''
\newblock {\em arXiv preprint arXiv:2406.00125}, 2024.

\end{thebibliography}

\end{document}